\documentclass[letterpaper, 10 pt, conference]{ieeeconf}

\IEEEoverridecommandlockouts

\usepackage{etextools}
\usepackage{amssymb}
\usepackage{float}
\usepackage{makeidx}
\usepackage{amsmath}
\usepackage{amsfonts}
\usepackage{bbm}
\usepackage{epsfig}
\usepackage{epsf}
\usepackage{dsfont}
\usepackage{psfrag}
\usepackage{verbatim}
\usepackage{multirow}
\usepackage{tabularx}
\usepackage{booktabs}
\usepackage[tight,footnotesize]{subfigure}
\usepackage{array}
\usepackage{soul}
\usepackage{footnote}
\usepackage[nocompress]{cite}

\usepackage{dblfloatfix}

\makeatletter
\let\NAT@parse\undefined
\makeatother

\usepackage{color, colortbl}
\definecolor{orange}{RGB}{255,128,0}
\usepackage[colorlinks,bookmarksnumbered,colorlinks=true,citecolor=orange,urlcolor=orange,linkcolor=orange,filecolor=orange]{hyperref}

\usepackage{graphicx}
\graphicspath{{Figures}}
\DeclareGraphicsExtensions{.pdf,.png}
\usepackage{bigstrut}
\usepackage[english]{babel}

\usepackage{csquotes}

\usepackage[T1]{fontenc}
\usepackage{algorithm, setspace}
\usepackage{algpseudocode}
\usepackage{url}

\newcommand{\p}[1]{\smallskip \noindent \textbf{{#1}.}}
\newcommand{\eq}[1]{Equation~(\ref{eq:#1})}

\usepackage{balance}

\newcommand{\sub}{\tau}

\title{\textbf{\LARGE In View or Out of View? A Vision-Aware Filter \\ that Infers and Reacts to the Human's Field of View}}

\author{Mengxue Fu$^1$, Ya-Chuan Hsu$^1$, Yinlong Dai$^2$, Erdem Bıyık$^{1\dagger}$, Dylan P. Losey$^{2\dagger}$
\thanks{$^\dagger$Equal advising. Research performed jointly by the $^1$\href{https://liralab.usc.edu/}{LiraLab}, Thomas Lord Department of Computer Science, University of Southern California, Los Angeles, CA 90089, and $^2$Collaborative Robotics Lab (\href{https://collab.me.vt.edu/}{Collab}), Department of Mechanical Engineering, Virginia Tech, Blacksburg, VA 24061.\newline Corresponding author's email: \texttt{mishafu@usc.edu}}
}

\begin{document}
\maketitle

\begin{abstract}
During human-robot interaction, sight plays an important role.
When the human sees a robot, they can observe the robot's actions and better predict the robot's future behaviors.
Conversely, when the human cannot see the robot, they become uncertain about the robot's intent and plans.
In this paper, we focus on the human's \emph{field of view}.
Existing human-aware planners either ignore this field of view and assume the human can observe the entire scene, or intentionally design their interactions so the robot is always visible to the human.
By contrast, we enable robots to infer what the human has and has not seen during interaction, i.e., the human's field of view.
We then apply this estimate as a \emph{vision-aware filter} on the robot's actions.
Given a nominal robot policy and a model of the interaction, we roll out the policy over a short horizon to see how the robot's state updates.
We then reweight the actions based on how those actions impact the robot's predicted visibility: higher weights are assigned to actions that bring the robot into (or away from) the human's line of sight.
In practice, our framework provides an online estimation and filtering process that repeatedly updates its understanding of the human's field of view and then locally perturbs robot actions based on the human's vision. 
We evaluate our vision-aware filter on robots that assist humans in collaborative cooking and coordinate with them in driving environments.
Across simulations and a 12-participant user study, we find that our vision-aware filter changes both performance and preference: participants completed over three times as many subtasks with in-view assistance while cooking, and had fewer collisions with cars that purposely stayed out of their view while driving.
\end{abstract}

\section{Introduction}

\begin{figure*}[t]
    \centering
    \includegraphics[width=1\linewidth]{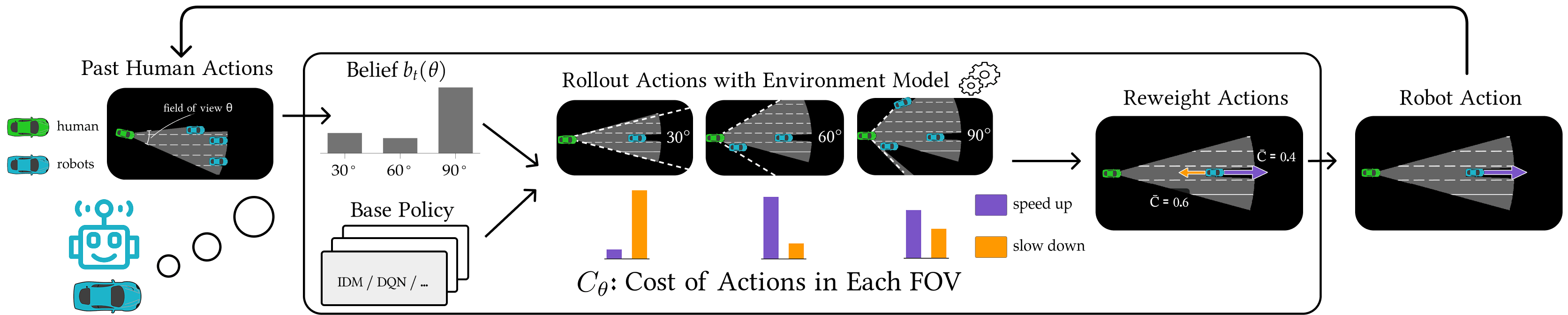}
    \caption{Overview of our vision-aware filter. The human (green) drives with an unknown field of view $\theta$ while robots (blue) control the surrounding traffic. Based on the human's actions the robots infer the human's current field of view (FOV). More formally, the robots maintain a belief $b_t(\theta)$ over what the human can see. The robots then use this belief to modify the actions of a base policy. The robot's base policy proposes candidate actions (e.g., speed up, slow down), and the robot leverages its model of the human and environment to roll out each candidate action and predict the time the robot will spend in the human's vision under each FOV hypothesis ($C_\theta$). The scores are weighted by the belief ($\bar C$). The robot then applies these scores to adjust the action distribution of the base policy, modifying how the robot acts to account for the human's FOV. Designers can tune whether this steers the robot into or away from the human's line of sight. Here our approach is used to minimize visibility, driving other robots out of the human's FOV.}
    \label{fig:method}
\end{figure*}

We focus on robots that must coordinate with humans whose field of view (FOV) is limited~\cite{howard1995binocular}.
Consider the example in Figure~\ref{fig:method}.
Autonomous cars are sharing the road with a human driver, and --- like any human driver --- the user can only see out of the windshield.
If the autonomous cars merge into lanes that are visible to the human, then the human must reason about these vehicles when driving (e.g., the human might slow down to pass the other cars more carefully).
But if the autonomous cars recognize what the human can and cannot see, then they can intentionally move out of the human's FOV and enable a more seamless driving experience for the user.
More broadly, FOV affects driving decisions when merging, yielding, and overtaking.

Existing works on human-robot collaboration recognize that modeling the human and predicting their behaviors are essential for seamless interactions \cite{carroll2019,wu2021too,kwon2020humans,nikolaidis2015efficient}.
But they typically make these predictions while \textit{assuming that the environment is fully visible to the human} \cite{carroll2019, xie2021learning, parekh2023learning}.
This is problematic from both human and robot perspectives.
When the robot fails to account for the human's line of sight, that robot may mistakenly attribute the human's decisions to the wrong intent, or incorrectly predict the human's future actions.
Accordingly, recent works have tried to address these issues by (a) assuming the human has a known, static field of view \cite{baker2017, kolb2024} and by (b) treating entry into that FOV as the way to assist \cite{hsu2025}. 
Here, two practical challenges emerge.
First, the human's FOV may change dynamically between interactions as they gain or lose focus, and second, we do not always want the robot to remain within the human's vision.

In this paper, we therefore explore how robots can dynamically infer and leverage the human's FOV during coordination.
Our insight is that:
\begin{center} \vspace{-0.3em}
    \textit{When robots infer the human's FOV, they can modify their actions to remain within (or stay outside of) that FOV without changing the robot's underlying policy.}
    \vspace{-0.3em}
\end{center}
Applying this insight, we develop a \textit{vision-aware filter} that can be connected to arbitrary policies for human-robot interaction.
The filter only requires (a) a predictive model of the human and robot, and (b) a designer-specified parameter that tunes whether the robot steers within or outside of the human's sight.
Given the human model, our approach leverages Bayesian inference to estimate the human's FOV based on their recent state-action pairs.
We then roll out future robot behaviors under the nominal policy, and evaluate how those behaviors overlap with the human's estimated FOV.
The policy actions are then reweighted based on whether the designer wants to steer into or out of sight. 
This intuitively results in a robot that locally modifies its behaviors based on the human's vision.
We emphasize that --- even in coordination tasks --- it is not necessarily beneficial to remain inside the human's FOV, and we explore this hypothesis by testing our method while biasing actions both towards and away from the human's sight.

Overall, we make the following contributions:

\p{Formulating Dynamic FOV Inference}
We enable robots to infer the human's FOV based on the system states and human behaviors.
We find that this FOV inference is tightly coupled to inferring the human's intended task, and that methods that fail to jointly consider both vision and task may misattribute the human's actions to the wrong cause.

\p{Developing a Vision-Aware Filter}
Equipped with our estimate of the human's FOV and intent, we next develop a model-based filter that locally adjusts the robot's actions to steer into or out of the human's line of sight.
This filter is invariant to the policy it is connected to, and therefore our approach generalizes across (and can be combined with) existing strategies for human-robot interaction.

\p{Comparing In-View and Out-of-View Coordination}
We conduct simulations and a user study where we compare robots that ignore and reason over the human's FOV.
Across a discrete collaborative cooking game and a continuous driving environment, we see that vision-aware robots are more effective and preferred partners.
We also show that moving into the human's FOV is not always beneficial, even when the robot is trying to help the human.
\section{Related Work} \label{sec:related}

\subsection{Assistive Robot Behavior and the Human's View}
Some existing work assumes that the robot's actions are in the human's view, and studies how robots should move given that a human is watching.
Within this context \textit{legible motion} shapes the robot's trajectory so that the human can quickly infer its intent~\cite{dragan2013legibility}, and information-revealing behaviors can convey facts the 
human is missing~\cite{shvo2022resolving, hiranaka2026fix, habibian2025survey}. 
On the other hand, other assistive actions do not depend on being seen. 
For example, anticipatory actions and subtask allocation improve teamwork~\cite{hoffman2007effects, habibian2022encouraging}, and conservative trajectories steer the robot away from the human to maintain safety and build trust~\cite{dautenhahn2006may, chen2020trust}.
However, prior works rarely consider how the human's FOV may evolve during the task --- meaning that these assistive actions fail to consider whether or not the human is currently watching.

\subsection{Models of What the Human Perceives}
Whether an action is seen depends on what the human perceives.
Theory-of-mind and interactive optimization-based formulations maintain a belief over the human's beliefs and goals~\cite{koster2013theory, gmytrasiewicz2005}, and situation awareness models track which parts of the environment the human has registered~\cite{senaratne2025framework}. 
More specifically, the human's FOV has been considered at several levels of fidelity. 
Some works include the human's heading direction or co-presence as a state feature~\cite{sartor2026epistemic} or model the FOV as a geometric cone that determines what the human can observe~\cite{hsu2025}. 
Occlusion-aware planning reasons about what is hidden from view and adjusts the robot's trajectory to compensate~\cite{zhang2021safe}. 
We note that these works generally treat the human's FOV as a constant, and decide whether or not to enter the human's FOV based on a task-specific heuristic.

\subsection{Human Factors in Visible Assistance}
Whether entering the human's view is desirable depends on the person and the task. 
Task complexity changes how much information a person can take in~\cite{sweller1988cognitive, wickens2008multiple}, and interruption studies show that poorly timed information is costly even when it is correct~\cite{cini2021relevance, hsu2026timing}. 
Preference also varies with expertise. 
Novices benefit from slow and steady disclosure, while experts prefer to be left to their own plan~\cite{carreno2022joint}.
Trust also builds gradually as people interact with the robot~\cite{campagna2025systematic}, so keeping the robot in view for longer may help build the user’s confidence.
The value of visible assistance therefore varies with load, task, and person.
When existing planners account for visibility at all, they treat it as something to increase, moving assistance into the human's view whenever the task reward permits it, and the opposite direction is left unexplored.
To the best of our knowledge, the literature lacks a controlled study that isolates this visibility variable by holding the task constant and evaluating in-view versus out-of-view assistance.
\section{Problem Statement} \label{sec:prob_statement}

We consider a two-agent human-robot setting.
Let $s = (s_R, s_H, s_W)$ be the system state, where the components represent the robot, the human, and the environment, and $s_H$ includes the human's position and orientation.
The robot fully observes state $s$ at every timestep.
The human has a limited FOV and only perceives $o = \phi(s, \theta)$.
We parameterize the human's FOV $\theta$ as the viewing angle, which follows the human's current heading, so the visible region changes as the human moves and turns (e.g., the cone shown in Figure~\ref{fig:method}).
The function $\phi$ retains the components of $s$ inside the FOV defined by $\theta$, and masks the rest.
We assume $\theta$ is fixed over an interaction and unknown to the robot.

The human and robot interact in a shared environment.
We assume that the human is performing some task with goal $g_H$ (e.g., completing a recipe or reaching a destination).
The human takes actions $a_H$ to complete this goal based on their history of observations, i.e., $a_H^t \sim \pi_H(\cdot \mid o^{0:t}, g_H)$.
We note that the human's behavior depends on $\theta$ via observations $o$.
The robot has a model of $\pi_H$, and takes action $a_R \in \mathcal{A}_R$ which transitions the state according to the shared dynamics $s^{t+1} = f(s^t, a_R^t, a_H^t)$.
Every robot action lands inside or outside the human's view, and the question we study is how this FOV placement affects human-robot coordination.

\section{Vision-Aware Filter} \label{sec:method}

Our method for inferring and accounting for the human's FOV consists of two parts, as shown in Figure~\ref{fig:method}.
First, the robot \textit{infers} the human's FOV $\theta$ during the interaction.
Since the human acts based on what they have observed, their actions carry evidence about $\theta$, and the robot can therefore leverage the human's actions $a_H$ to maintain and update a posterior over $\theta$.
Second, the robot \textit{filters} its own actions to account for the human's FOV.
We perform this vision-aware filtering in a policy-invariant way by refining the action distribution of a given base policy to bias those actions towards (or away from) the human's FOV.
More formally, we add a cost term that the designer can tune to shape this filter.
The sign of the cost determines whether the robot's preferred coordination strategy is to steer into or out of the human's view, which lets us treat the direction of steering as a controlled variable and evaluate both directions in Sections~\ref{sec:sim} and~\ref{sec:study}.
We note that our method relies on having a model of the system, including both the human's policy $\pi_H$ and the environment dynamics $f$.

\subsection{Human Model} \label{sec:human_model}

The robot first needs to build an internal human model in order to (a) infer the human's FOV from observed human behavior and (b) predict the human's next actions using the FOV inference. 
The human's true policy is unknown to the robot. 
Instead, the robot has an approximate human model, represented as a conditional distribution over human actions:
\begin{equation} \label{eq:M1}
    \hat\pi_H(a_H^t \mid h^t,\theta) := \hat P\left(a_H^t \mid s^{0:t}, a_H^{0:t-1}, a_R^{0:t-1}, \theta\right)
\end{equation}
where $\hat P$ approximates the true human policy $\pi_H$, $s^{0:t}$ is the state history observed by the robot, $a_H^{0:t}$ denotes the human actions observed up to timestep $t$, and $h^t = (s^{0:t},a_H^{0:t-1},a_R^{0:t-1})$ denotes the interaction history.
Intuitively, the human model describes how likely the human is to choose action $a_H$ at time $t$ based on the interaction history and what they have observed.
By connecting observed human actions to their hidden FOV $\theta$, this model enables the robot to infer the human's FOV as shown in Section~\ref{sec:infer-fov}.
The human model may be learned (for example using a neural network trained on recent human behaviors).
In our implementation, we approximate the human's policy through a series of predefined decision rules.

\subsection{Inferring Human's Field of View} \label{sec:infer-fov}

We use the approximate human model $\hat\pi_H$ to infer the human's FOV from observed actions.
The robot maintains a posterior belief:
\begin{equation} \label{eq:M2}
    b_t(\theta) = P\!\left(\theta \mid h^t, a_H^t \right)
\end{equation}
This belief is updated recursively according to how likely the observed human action is under each candidate $\theta$, as predicted by the human model.
Using Bayes' theorem:
\begin{equation} \label{eq:M3}
    b_t(\theta)
    \propto
    \hat\pi_H(a_H^t \mid h^t,\theta)
    b_{t-1}(\theta)
\end{equation}
The likelihood measures how well a candidate $\theta$ explains the human's observed action $a_H^t$; that is, it captures what the robot thinks the human can see based on how they act. 
Utilizing this belief, we next construct an FOV filter that can be applied to any robot base policy to steer its actions for better coordination with the human.

\subsection{Vision-Aware Policy Refinement} \label{sec:M3}

We next use the posterior belief $b_t(\theta)$ to refine the action distribution proposed by a nominal robot policy.

\p{Base Policy}
Let $\pi_R$ be a base robot policy that maps the observation history to a distribution over robot actions:
\begin{equation}
    a_R^{t} \sim \pi_R(\cdot \mid s^{0:t}, a_H^{0:t-1})
\end{equation}
Note that we impose no restrictions on the form of this nominal robot policy. 
It may or may not use the human model $\hat\pi_H$, and it does not need to directly account for the inferred human's FOV $b_t(\theta)$.
We treat the base policy as a black box and do not assume it to be optimal. 
Its action distribution may contain errors arising from an inaccurate human model, imperfect task reasoning, or other modeling approximations. 
However, we do assume that any safety requirements of the domain are enforced at this level, namely, the base policy assigns zero probability to unsafe actions. 

\p{Vision-Aware Filter} 
We modify the outputs of the base policy using our vision-aware filter.
Given the base action distribution at timestep $t$ and the posterior belief $b_t(\theta)$, let $C_\theta(s^t,a_R)$ denote the predicted cost of taking action $a_R$ at $s^t$ when the human has FOV $\theta$. 
We will describe this cost $C$ further in the following section.
For now, we aggregate the costs across $\theta$ hypotheses according to:
\begin{equation}
    \bar{C}_t(s^t,a_R) = \sum_{\theta\in\Theta}
    b_t(\theta)\,
    C_\theta(s^t,a_R).
    \label{eq:fov_cost}
\end{equation}
The FOV-aware action is then selected as
\begin{multline}
    a_R^t \in \arg\!\min_{a_R\in\mathcal{A}_R}
    \Big[
        -\log\pi_R(a_R \mid s^{0:t}, a_H^{0:t-1}) \\ + \lambda\,\bar{C}_t(s^t,a_R)
    \Big]
    \label{eq:fov_filtered_action}
\end{multline}
where $\mathcal{A}_R$ is the robot's action space. 
The first term favors actions assigned high probability by the base policy, while the second term incorporates the FOV-conditioned rollout cost. 
The coefficient $\lambda$ controls the influence of the FOV filter and hence the robot's behavior: when $\lambda=0$, the method reduces to the base policy, whereas increasing the magnitude of $\lambda$ gives greater weight to the FOV-conditioned evaluation.

\p{FOV-Conditioned Cost}
The choice of the FOV-conditioned cost $C_\theta$ is open-ended. 
To preserve generality, its construction should not depend on domain- or task-specific knowledge. 
In this work, we use a visibility-based cost. 
We first assume access to the transition model $f$ that predicts the next state given the current state and both agents' actions (see Section~\ref{sec:prob_statement}).
Let $v(s,\theta)\in\{0,1\}$ indicate whether the robot lies inside the visible region induced by $\theta$ in state $s$. 
For each candidate action $a_R$, we use $f$ to generate an $H$-step rollout, with the robot executing $a_R$ initially and following its base policy thereafter, while the human acts according to the approximated human model $\hat{\pi}_H$.
The cost is the expected fraction of the rollout the robot spends in the human's view:
\begin{equation}
    C_\theta(s^t,a_R) = \frac{1}{H}\sum_{k=1}^{H}
    \mathbb{E}\!\left[v(s^{t+k},\theta)\right].
    \label{eq:visibility_cost}
\end{equation}
With this choice, $\lambda>0$ in Equation~\eqref{eq:fov_filtered_action} penalizes visibility and drives the robot out of the human's FOV, $\lambda<0$ rewards visibility and drives the robot into the human's FOV, and $\lambda=0$ recovers the base policy.

\p{Task and FOV Trade-Off}
Since the FOV filter does not use domain- or task-specific knowledge, its adjustments to the robot's actions may conflict with the base policy's task objective.
While the filter can potentially reweight actions to favor lower visibility costs and sacrifice task performance, we impose three constraints that protect the base policy's performance.

First, the filter never selects an action to which the base policy assigns zero probability. 
In practice, the filter reweights the base policy's top-$k$ actions. 
Second, each candidate must remain within a bounded margin of the base policy's own choice.
More specifically, when the base policy plans a sequence of actions toward its current objective, the length of that sequence serves as a cost $d$, and a candidate is discarded if its own cost exceeds $(1+c)\cdot d$, where $c \geq 0$ is a hyperparameter. 
For base policies that emit a single action rather than a distribution, candidates are bounded offsets on that action, penalized in proportion to their deviation. 
Finally, if no candidate satisfies these constraints, or if the posterior $b_t(\theta)$ has not yet concentrated, the filter returns the base policy's action unchanged. 
The filter thus intervenes only within the base policy's own tolerance, in the spirit of safety filters for safety-critical control~\cite{hsu2023safety}.

\section{Simulation Experiments} \label{sec:sim}

With our vision-aware filter introduced, we next want to test whether reweighting an FOV-agnostic base policy by visibility alone can improve coordination between agents.
We start with simulated environments where we can directly control the human model and test a variety of base robot policies.
We consider two types of collaborative domains (overcooked and driving), and explore whether steering towards or away from the human's FOV affects coordination differently in these different domains.

\subsection{Experimental Domains} 
We evaluate our vision-aware filter in two domains that require different types of coordination, allow diverse, distinguishable FOV configurations, and span discrete and continuous state and action spaces.

\begin{figure*}[t]
    \centering
    \includegraphics[width=\linewidth, height=1.15in]{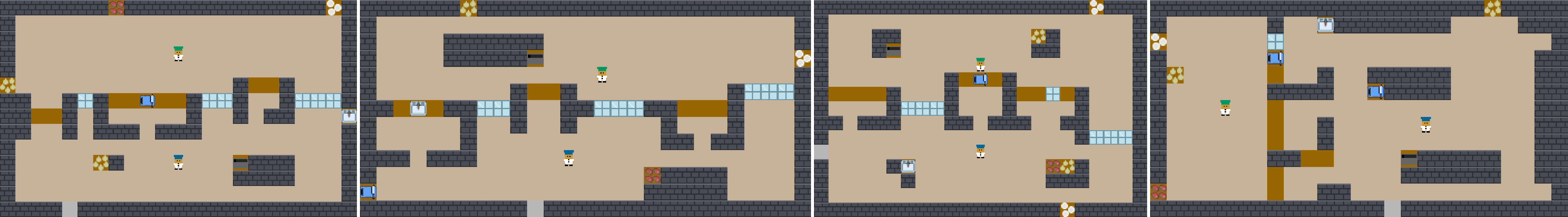}
    \caption{Overcooked environment. Here we show the four kitchen layouts used in our simulation experiments. We visualize the start state for each option. From left to right: back bar, banquet pass, chefs table, and divider. Brick tiles are walls, which block the human's line of sight; pane tiles are glass, which is impassable but see-through. Each kitchen contains meat, onion, and plate dispensers, a grill, a chopping board, a sink, and a serving window. The green-hat agent is the robot and the blue-hat agent is the human.}
    \label{fig:layouts}
\end{figure*}

\begin{itemize}
\item \textbf{Overcooked} (Figure~\ref{fig:layouts}) is a collaborative kitchen environment \cite{hsu2025} built on top of Overcooked-AI by \cite{carroll2019}.
A human and robot must collaborate and deliver as many dishes as possible within a limited number of timesteps. 
The human sees only within a cone centered on its heading, and walls further occlude its line of sight. The four layouts confine the agents to separate sections of the kitchen, enforcing coordination by requiring agents to pass ingredients to each other to complete orders. \vspace{0.25em}

\item \textbf{Highway} (Figure~\ref{fig:method}) is a continuous, safety-critical driving environment \cite{highway-env} where the human operates an ego vehicle through dense traffic controlled by robot policies. 
We customize the environment to use an 8-lane road with 80 vehicles at $1.5\times$ the density of the environment's defaults on the ``highway-v0'' scene, making interactions with surrounding traffic more frequent.
The human's goal is to travel a fixed distance as quickly as possible while avoiding collision with surrounding vehicles.
The robots do not share the human's task; they influence the human only by shaping how the other vehicles share the road.
\end{itemize}

\subsection{Simulated Humans}
In both domains the simulated human has a limited FOV.
This human acts greedily to make immediate progress towards the task's objective following the human model in Section~\ref{sec:human_model}. 
We next introduce the simulated human's observation and action spaces in each domain.

In Overcooked, the simulated human's observation space includes ingredient locations currently within its FOV and those retained in a memory buffer from the past $40$ timesteps. Visible locations are updated to their observed values, while remembered locations expire $40$ timesteps after they were last observed.
Memory is essential for a plausible partner, since a human who instantly forgets everything outside its FOV would be ineffective, while one who never forgets would be unrealistic. 
Following prior work \cite{wu2021too}, we define a human's high-level action space as a set of subtasks $\sub \in \mathbb{T}$, allowing the human to plan at the subtask level. 
Each subtask has preconditions under which it is available. 
For example, ``cook food'' requires that uncooked food is present and the grill is empty. 
At each step, the human selects the highest-priority subtask available according to memory, and plans to execute it with A*, treating locations it has never observed as unwalkable. 
If it last saw the robot carrying an item it needs, it goes to look for where that item was placed. 

In Highway, the human's observation space consists of information about surrounding vehicles within their FOV and a sensing distance of $40\,\mathrm{m}$. 
The human's action space consists of speed adjustments between $1$-$35\,\mathrm{m/s}$, as well as moving to adjacent lanes.
The simulated human changes lanes when they believe a faster lane is free, unless the merging action leads to a conflict with a visible vehicle.

\subsection{Base Policies}
For each domain we implement a distinct set of base robot policies. 
Viewed together, we test five policies ranging from rule-based to learned control, and from no partner reasoning to full Bayesian inference.
None of these nominal policies reason over the human's FOV.
We test whether our filter is able to synergize with these different base policies and improve team coordination.
In \textbf{Overcooked}, we consider:
\begin{itemize}
    \item \textit{Greedy} \cite{carroll2019}, which selects subtasks by task progress and ignores the presence of a partner;
    \item \textit{Solo}, which adjusts the robot's behavior based on the human's current position to minimize interference;
    \item \textit{Bayes} \cite{wu2021too}, which collaborates and infers the human's current subtask while assuming that the human observes the full system state.
\end{itemize}
Each policy outputs its subtask ranking to the filter as a discrete action distribution. In \textbf{Highway}, we consider:
\begin{itemize}
    \item \textit{IDM}, where each background vehicle follows the Intelligent Driver Model \cite{treiber2000idm} for car-following with MOBIL lane changes \cite{treiber2016mobil};
    \item \textit{DQN}, a pretrained deep Q-network policy that selects among discrete acceleration and lane-change actions.
\end{itemize}

\subsection{Filter Design}
In both domains the vision-aware filter follows Section~\ref{sec:method}.
The filter is given candidate actions from the base policy, forward simulates each candidate action under the human and environment models, and then scores that action by its expected visibility under the estimated human FOV.
Finally, the filter chooses the candidate (or locally modifies the candidate) with the lowest cost according to \eq{fov_filtered_action}.

In Overcooked, candidates are the base policy's three highest-ranked subtasks; rollouts that are used to compute the FOV-conditioned cost extend $H=12$ timesteps, roughly the duration of one subtask. 
In Highway, the filter runs on every neighboring vehicle within $45\,\mathrm{m}$ of the human, and adjusts only each vehicle's acceleration. 
Lane-change decisions are always left to the base policy. 
Candidates are bounded offsets on the nominal acceleration. 
To score a candidate, the filter rolls the environment forward for $3\,\mathrm{s}$ under that acceleration, with neighboring vehicles at constant velocity. 
  
\subsection{Evaluation}
In Overcooked we evaluate all combinations of the four layouts (see Figure~\ref{fig:layouts}), four FOVs $\theta \in \{30^\circ, 60^\circ, 90^\circ, 360^\circ\}$, and three base policies, with 20 seeds per combination. 
Each episode lasts up to 400 timesteps and ends once three dishes have been delivered.
In Highway we evaluate all combinations of the two base policies and three FOVs, $\theta \in \{30^\circ, 60^\circ, 90^\circ\}$, with 20 seeds per combination.
Each session lasts 60 seconds of simulated driving; a collision restarts the traffic while elapsed time and distance carry over. 
We report mean dishes delivered (task success) in Overcooked and collisions (safety), distance covered (traffic efficiency), and the human's completed lane changes (maneuvering induced by the traffic) in Highway. 
For both domains we also report the filtered policy's total computation per timestep and its overhead relative to the base policy, which together determine whether this vision-aware filter can be applied in real-time and support the user study in Section~\ref{sec:study}.

\subsection{Results}
We compare three versions of our filter: no adjustment to the base policy ($\lambda = 0)$, steering the robot away from the human's inferred FOV ($\lambda > 0$), and steering the robot into the human's FOV ($\lambda < 0)$.
In our results we refer to these conditions as \textbf{Base}, \textbf{Out}, and \textbf{In}, respectively.

\p{Overcooked} Table~\ref{tab:overcooked_sim} reports mean dishes delivered per condition. 
The \textit{in-view filter} ($\lambda < 0)$ improves over every base policy at every FOV, increasing mean dishes per episode (averaged over layouts and FOVs) from 1.16 to 1.40 for greedy, 1.14 to 1.39 for solo, and 0.69 to 1.06 for Bayes. 
Note that even the omnidirectional $\theta=360^\circ$ setting falls short of full observability, since the layouts' walls still occlude the view, leaving room for the filter to improve. 

The gain is largest for Bayes, whose assumption that the human observes the full state is most wrong at narrow FOVs, and which the filter helps to correct. 
The benefit is steepest precisely where sight is narrowest: at $\theta = 30^\circ$ the in-view filter increases deliveries by over 60\% relative to each base policy, and the relative gain is smaller at wider FOVs. 
In general, the less the human can see, the more valuable it is for the robot to work where the human is looking.

\begin{table}[t]
\centering
\caption{Overcooked simulation: mean dishes delivered per 400-timestep episode (maximum 3 deliverable orders per episode). Results are averaged over four layouts and 20 seeds. Each row corresponds to a different human FOV $\theta$. Columns correspond to the vision-aware filter: unfiltered (base policy), out-of-view filter (out, $\lambda > 0$), in-view filter (in, $\lambda < 0$); best is in bold.}
\label{tab:overcooked_sim}
\small
\setlength{\tabcolsep}{4pt}
\begin{tabular}{l ccc ccc ccc}
\toprule
 & \multicolumn{3}{c}{Greedy} & \multicolumn{3}{c}{Solo} & \multicolumn{3}{c}{Bayes} \\
\cmidrule(lr){2-4} \cmidrule(lr){5-7} \cmidrule(lr){8-10}
$\theta$ & Base & Out & In & Base & Out & In & Base & Out & In \\
\midrule
$30^\circ$  & 0.43 & 0.08 & \textbf{0.70} & 0.40 & 0.08 & \textbf{0.65} & 0.20 & 0.12 & \textbf{0.50} \\
$60^\circ$  & 1.27 & 1.25 & \textbf{1.48} & 1.20 & 1.25 & \textbf{1.45} & 0.80 & 1.05 & \textbf{1.10} \\
$90^\circ$  & 1.25 & 1.30 & \textbf{1.50} & 1.42 & 1.28 & \textbf{1.50} & 0.73 & 1.02 & \textbf{1.12} \\
$360^\circ$ & 1.70 & 1.70 & \textbf{1.93} & 1.55 & 1.70 & \textbf{1.95} & 1.02 & 1.45 & \textbf{1.50} \\
\bottomrule
\end{tabular}
\end{table}

\p{Highway} Table~\ref{tab:highway_sim} reports collisions and traveled distance per session. 
Results show that staying \textit{out of the human's view} ($\lambda > 0$) is the better approach in this domain, helping under both base policies at all but the narrowest FOV under IDM. 
Under DQN traffic, the out-of-view filter reduces collisions at every FOV, from 4.03 to 2.62 per session on average. 
Under IDM traffic, it reduces collisions at $\theta = 60^\circ$ and $90^\circ$ while adding roughly 180\,m of distance covered per session. 
We attribute this to how the filter reshapes traffic, namely, vehicles ahead of the human accelerate out of the FOV, traffic spreads out, and the human brakes less. 

The human's maneuvering shows the same effect. 
Under IDM traffic, the in-view filter pulls vehicles into the FOV and raises the human's lane changes from 19.9 to 27.8 per session, alongside more collisions. 
Under the DQN traffic, the out-of-view filter reduces the number of lane changes from 13.3 to 10.0 as vehicles clear out of the human's path. 
Neither direction helps under IDM at $\theta = 30^\circ$. 
We attribute this to the FOV itself rather than to FOV inference: with such a narrow view, the human drives more recklessly, colliding mostly with vehicles it never gets to see, so rearranging which vehicles are visible cannot fully overcome the natural FOV limitation.

\begin{table}[t]
\centering
\caption{Highway simulation: mean collisions\,$|$\,distance (m) per 60\,s session over 20 paired seeds. Each row corresponds to a different human field of view $\theta$. Columns correspond to the different traffic policies and vision-aware filters: unfiltered (base), out-of-view filter (out), in-view filter (in); best in bold (fewer collisions; ties broken by more distance).}
\label{tab:highway_sim}
\small
\setlength{\tabcolsep}{3pt}
\resizebox{\linewidth}{!}{%
\begin{tabular}{l ccc ccc}
\toprule
 & \multicolumn{3}{c}{IDM} & \multicolumn{3}{c}{DQN} \\
\cmidrule(lr){2-4} \cmidrule(lr){5-7}
$\theta$ & Base & Out & In & Base & Out & In \\
\midrule
$30^\circ$ & \textbf{3.50}\,$|$\,1778 & 4.20\,$|$\,1900 & 3.90\,$|$\,1711 & 5.15\,$|$\,1437 & \textbf{2.85}\,$|$\,1461 & 4.10\,$|$\,1284 \\
$60^\circ$ & 0.75\,$|$\,1686 & \textbf{0.40}\,$|$\,1907 & 1.50\,$|$\,1667 & 3.55\,$|$\,1268 & \textbf{2.50}\,$|$\,1232 & 3.05\,$|$\,1267 \\
$90^\circ$ & 0.50\,$|$\,1702 & \textbf{0.35}\,$|$\,1902 & 1.00\,$|$\,1699 & 3.40\,$|$\,1237 & \textbf{2.50}\,$|$\,1248 & 3.45\,$|$\,1372 \\
\bottomrule
\end{tabular}}
\end{table}

\p{Computation Time} 
The filter fits both domains' real-time budgets. 
In Overcooked the filter computes its action in 91\,ms on average, against under 1\,ms for the underlying Bayes policy alone. 
Although the filter adds substantial overhead relative to the base policy, the total remains well within the real-time budget of the game, which is around 3 Hz.
This added time is caused by rolling out candidate actions and considering their future effects on visibility: the time delay of the filter is correlated with the simulation time and depth $H$.
In Highway, computing all background vehicles' actions takes 21\,ms with the filter compared to 17.5\,ms with just the base policy under the IDM model, and 10--14\,ms for the DQN model. 
The total is still under the real-time budget for Highway, which is around 15 Hz. 
We therefore run the same filters without any changes in the user study.

\section{User Study} \label{sec:study}

\begin{figure}[t]
    \centering
    \includegraphics[width=\linewidth]{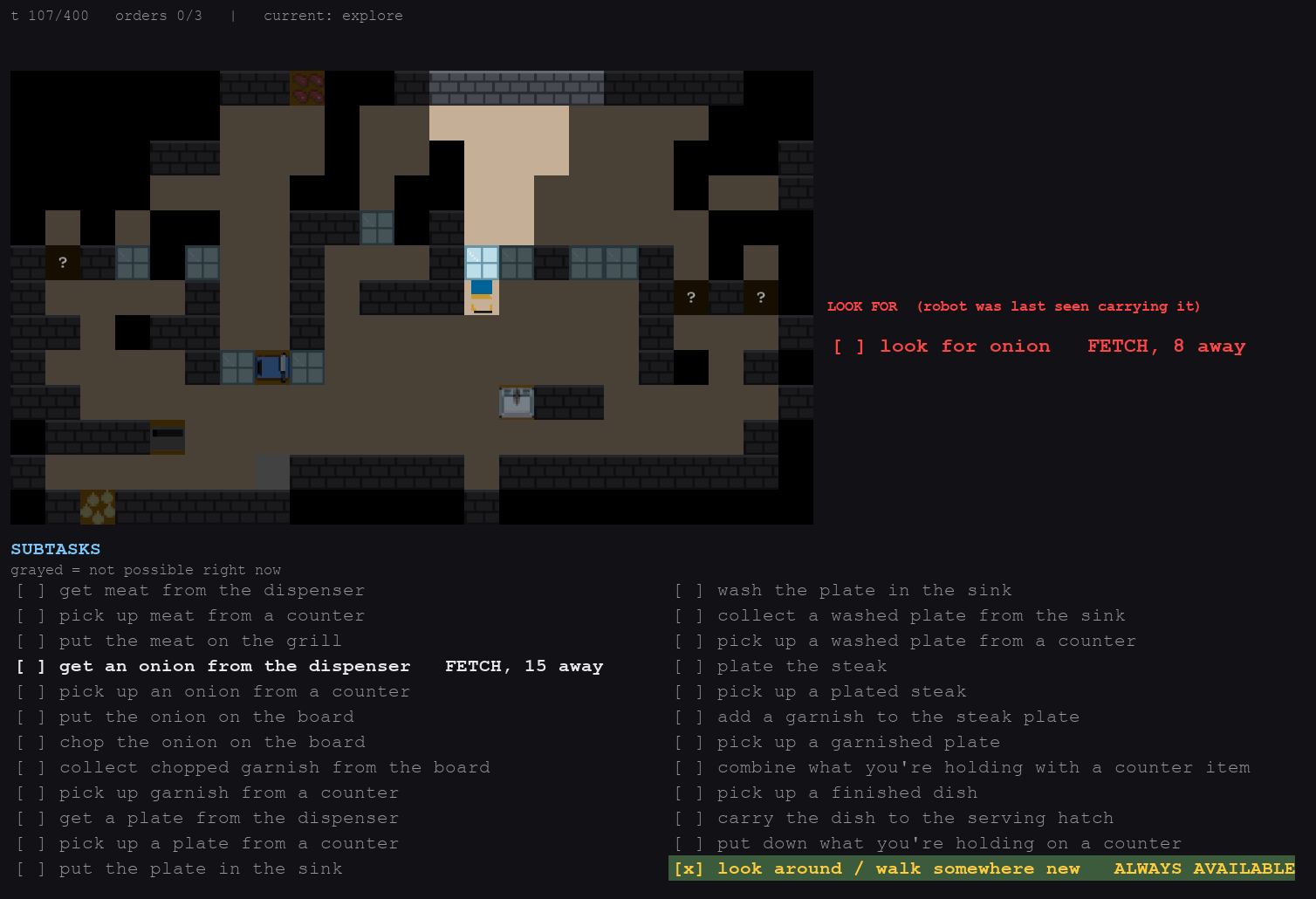}\\[6pt]
    \includegraphics[width=\linewidth]{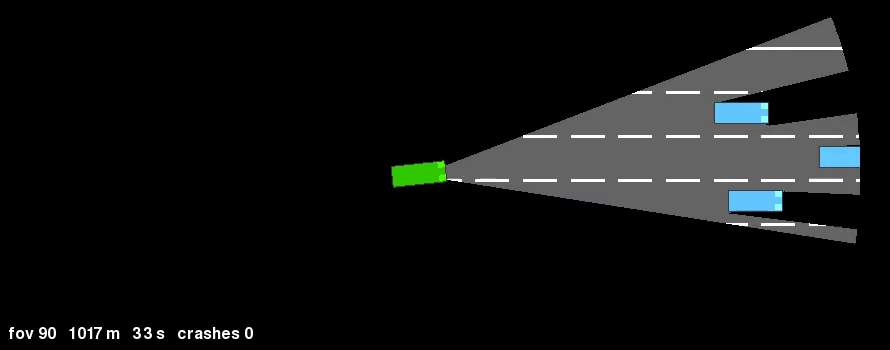}
    \caption{The user study interfaces. Top: Overcooked, where the participant sees only the region inside their FOV and selects among currently available high-level subtasks from the menu. Bottom: Highway, where the participant steers the ego vehicle and sees only vehicles inside their FOV.}
    \label{fig:study_interfaces}
\end{figure}

Our simulations enabled us to test the vision-aware filter in controlled conditions.
Next, we explore how this filter helps robots collaborate with \textit{actual humans}.
Our user study setup follows the design of our simulations, again using the \textbf{Overcooked} and \textbf{Highway} domains where a robot and human must collaborate.
We apply visual masks to these environments to regulate what the human can and cannot perceive (see Figure~\ref{fig:study_interfaces}). 
The study tests whether the effects observed in simulation hold with real users, and additionally examines how users perceive the differences between the base policy and the in-view and out-of-view filters.

\p{Participants}
We conducted the study with 12 participants (3 female, 9 male; ages 20--63, mean $= 27.9$). 
Nine of our participants had experience playing the commercial game Overcooked. 
One participant played driving simulators regularly, and seven participants had used driving simulators occasionally. 
Four participants had prior research experience with the Overcooked-AI environment \cite{carroll2019}, and three with the Highway environment \cite{highway-env}.
All participants provided informed written consent following the IRB protocol at the University of Southern California.

\p{Experimental Setup}
Each participant interacted with three types of robot agents: the base policy alone (no filter, $\lambda = 0$), the base policy with the out-of-view filter ($\lambda > 0$), and the base policy with the in-view filter ($\lambda < 0$).
The order of robot policies was counterbalanced across participants, and conditions were unnamed: participants knew only that they would interact with three different robot policies.
The FOV was assigned between users ($\theta = 30^\circ$ for 6 participants, $60^\circ$ for 6) and held fixed for that user in both domains. 
These two $\theta$ values were chosen based on preliminary experiments; we are most interested in how the filter operates in low-visibility settings.
Each participant completed the Overcooked and Highway domains in one session, beginning each domain with a familiarization round under a near-unrestricted view that was excluded from analysis.

In Overcooked the participant played one 400-timestep episode with each robot policy using \emph{Bayes} as the base policy.
As shown in Figure~\ref{fig:study_interfaces} (top), the participant saw the kitchen only inside their FOV and acted at the subtask level. 
A menu listed the high-level subtasks, and a row was clickable only when that subtask was currently available given what the participant could observe. 
Clicking a row made their chef carry that task out automatically, and the SPACE bar canceled the current subtask.
In Highway, the participant drove the ego vehicle with the arrow keys through IDM traffic for 60\,s (Figure~\ref{fig:study_interfaces}, bottom); a collision reset the surrounding traffic while elapsed time and distance carried over.
After each episode, participants rated the robot on 7-point Likert scales. 
After completing each domain, participants ranked the three robot policies and optionally described any differences they noticed.

\p{Objective Results} 
We measured the human's subtask completions and dishes delivered in Overcooked, and collisions and distance covered in Highway.
Our results are summarized in Table~\ref{tab:study_objective}.
In Overcooked --- with real humans in the loop --- the \textit{in-view} filter's advantage from simulation not only holds but widens. 
Participants completed $11.75$ subtasks on average with the in-view robot, over three times the $3.17$ they completed with the base policy ($p < .001$, Wilcoxon signed-rank test).
Six of the seven dishes delivered across the entire study came from the in-view condition. 
This affirms the simulation results, showing that acting within the human's FOV keeps their knowledge of the kitchen current: in the study, the subtask of looking for an item could only be selected once the participant observed the robot carrying that item, so a robot that passed through where the participant was looking directly expanded what they could act on, making the overall task more effective.

In Highway, as in simulation, participants performed best when the robots stayed \textit{out} of their FOV. 
Participants drove further ($1855$ versus $1779$\,m on average) with fewer collisions ($1.17$ versus $1.25$ per session) as compared to the base policy.
When the robot controlled the vehicles to stay out of the human's line of sight, these vehicles often sped up (moving away from the human), allowing the human to drive unimpeded.
By contrast, the in-view filter drew vehicles into the participant's limited sight, and then the participants had to navigate around these blocking vehicles.
Hence, filtering the robot's actions to remain within the human's vision resulted in the most collisions and the least distance covered ($1.42$ collisions, $1733$\,m). 
Together, the two domains show that a robot which infers the human's FOV can steer its actions either into or out of view, potentially improving the team's coordination.
But it is not as simple as always trying to maximize the robot's visibility --- how the robot should react to the human's FOV depends on the current domain.

\begin{table}[t]
\centering
\caption{User study objective results ($N = 12$): mean per participant with each robot policy, except dishes delivered, which are totaled across all participants. Columns correspond to the base policy (base), out-of-view filter (out, $\lambda > 0$), and in-view filter (in, $\lambda < 0$); best is in bold.}
\label{tab:study_objective}
\small
\setlength{\tabcolsep}{4pt}
\begin{tabular}{ll ccc}
\toprule
Domain & Metric & Base & Out & In \\
\midrule
\multirow{2}{*}{Overcooked} & Subtasks completed  & 3.17 & 5.67 & \textbf{11.75} \\
 & Dishes delivered (total)  & 0 & 1 & \textbf{6} \\
\midrule
\multirow{2}{*}{Highway} & Collisions per session  & 1.25 & \textbf{1.17} & 1.42 \\
 & Distance (m)  & 1779 & \textbf{1855} & 1733 \\
\bottomrule
\end{tabular}
\end{table}

\begin{figure*}[t]
    \centering
    \includegraphics[width=\linewidth]{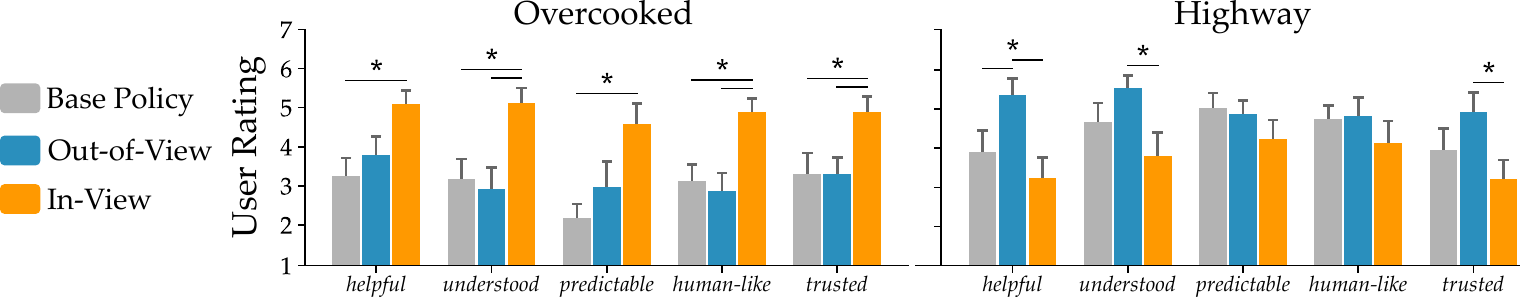}
    \caption{User study subjective ratings. Participants responded to a 7-point Likert scale survey after working with each filter type. Error bars show standard error, and an $\star$ denotes statistical significance ($p< .05$). We find that people preferred robots that used our vision-aware filter (the colored bars); the correct way to leverage that filter (in-view or out-of-view) depended on the domain.}
    \label{fig:likert}
\end{figure*}

\begin{figure}[t]
    \centering
    \includegraphics[width=\linewidth]{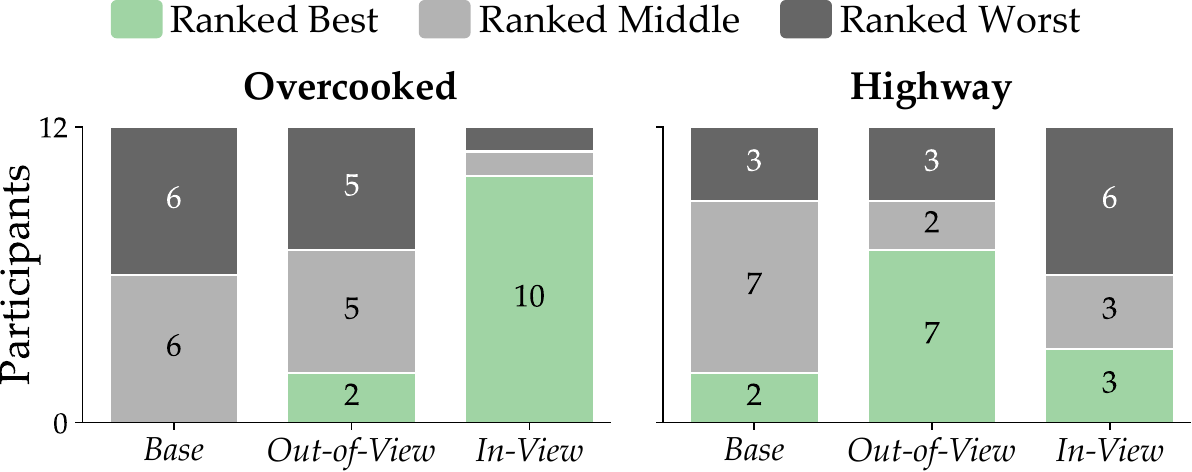}
    \caption{Forced choice comparison. Our 12 participants ranked the filter type from best to worst in both experimental domains. In Overcooked participants preferred \textit{in-view}, while in Highway participants preferred \textit{out-of-view}.}
    \label{fig:ranking}
\end{figure}

\p{Subjective Results}
After each interaction participants rated their agreement with statements on 7-point Likert scales, adapted from established fluency and trust questionnaires in human-robot collaboration \cite{hoffman2019fluency, jian2000trust}: ``the robot was helpful'' (\textit{helpfulness}), ``I understood what the robot was doing and why'' (\textit{understanding}), ``I could predict what the robot would do next'' (\textit{predictability}), ``the robot behaved the way a human would'' (\textit{human-likeness}), and ``I trusted the robot to do the right thing'' (\textit{trust}).

Figure~\ref{fig:likert} reports the user rating for these scales, and Figure~\ref{fig:ranking} reports the participants' forced rankings for each filter type.
In Overcooked the participants' preferences matched their performance: the in-view robot was rated above the base policy on all five constructs ($p < .05$ for all), and was ranked best by 10 of the 12 participants.
The other two participants ranked the out-of-view filter the best, potentially representing users who prefer completing tasks alone with as little interaction as possible with the robot. 
Participants not only benefited from visible assistance but also recognized it.
One participant wrote that the in-view robot ``\textit{passed the food to the other side of the counter [...] as if it was intentionally showing the player it is carrying the food.}''

In Highway the out-of-view robot policy was rated as significantly more helpful than the base policy ($p = .036$), with trust and understanding trending in the same direction ($p = .094$ for both).
A total of 7 of 12 participants selected the out-of-view policy as their preference. 
By contrast, the in-view robot was rated lowest on every construct: directing traffic into a limited view was not received as assistance but rather as interference.
Participants also provided descriptions, including one describing the out-of-view robots as ``\textit{actively mak[ing] way for the ego}'' while the in-view robot ``\textit{somehow seemed actively in [ego vehicle's] way.}'' 
Across both domains participants were not told which types of filters they were working with, yet their rankings and descriptions recovered the distinction the filter implements.

\section{Conclusion} \label{sec:conclusion}

In this paper we introduced a vision-aware filter that accounts for what the human can see during human-robot interaction.
Our filter had two main components: (a) dynamically updating its estimate of the human's current FOV, and then (b) modifying the actions of a base policy to steer the robot towards --- or away from --- the human's estimated FOV.
We achieved this action modification in a policy-invariant manner by reweighting the action distribution according to the predicted visibility of each candidate action.
Although our approach assumes access to a dynamics model and an estimated human model, the filter \textit{does not require} any task knowledge.
Instead, we provide a parameter that designers can tune to adjust whether our filter increases or decreases the robot's visibility. 
During controlled simulations and a user study we found that inferring and accounting for the human's FOV can significantly impact human-robot collaboration.
The right way to steer the robot, however, appears to depend on the task and domain.
In our collaborative cooking environment, maximizing the robot's visibility resulted in more efficient teamwork.
Conversely, in our shared driving environment, keeping other vehicles out of the human's line of sight provided a clearer road and enabled faster motions.

\p{Limitations}
While our current filter assumes the field of view is constant within an interaction, real attention shifts with focus and task, so extending the posterior to track a moving, task-dependent view is a natural next step. 
More broadly, the benefits of a purely visibility-based signal suggest that visibility could be folded into end-to-end learning, training robot policies that can decide for themselves whether and when to be seen. 
Our results strongly indicate that when robots make this choice well, they become better partners when coordinating with humans.

\section*{Acknowledgments}
The authors thank the USC Center for Advanced Research Computing for providing us with compute resources.

\balance
\bibliographystyle{IEEEtran}
\bibliography{references}
\end{document}